\documentclass[conference]{IEEEtran}
\IEEEoverridecommandlockouts
\usepackage{cite}
\usepackage{amsmath,amssymb,amsfonts}
\usepackage{algorithmic}
\usepackage{graphicx}
\usepackage{textcomp}
\usepackage{booktabs}
\usepackage{xcolor}
\def\BibTeX{{\rm B\kern-.05em{\sc i\kern-.025em b}\kern-.08em
    T\kern-.1667em\lower.7ex\hbox{E}\kern-.125emX}}
\makeatletter
\def\ps@IEEEtitlepagestyle{%
\def\@oddfoot{\mycopyrightnotice}%
\def\@evenfoot{}%
}
\def\mycopyrightnotice{%
{\footnotesize 978-1-6654-7189-3/22/\$31.00~\copyright~2022 IEEE\hfill}
\gdef\mycopyrightnotice{}
}
\begin{document}

\title{A Keypoint Based Enhancement Method for Audio Driven Free View Talking Head Synthesis\\
{
\footnotesize 
% \textsuperscript{*}Note: Sub-titles are not captured in Xplore and should not be used
}
}

\author{\IEEEauthorblockN{Yichen Han, Ya Li, Yingming Gao, Jinlong Xue}
\IEEEauthorblockA{\textit{School of Artificial Intelligence} \\
\textit{Beijing University of Posts and Telecommunications}\\
Beijing, China \\
adelacvgaoiro@bupt.edu.cn, yli01@bupt.edu.cn,\\ yingming.gao@outlook.com, jinlong\_xue@bupt.edu.cn}
% \and
% \IEEEauthorblockN{}
% \IEEEauthorblockA{\textit{School of Artificial Intelligence} \\
% \textit{Beijing University of Posts and Telecommunications}\\
% Beijing, China \\
% }
% \and
% \IEEEauthorblockN{}
% \IEEEauthorblockA{\textit{School of Artificial Intelligence} \\
% \textit{Beijing University of Posts and Telecommunications}\\
% Beijing, China \\
% }
% \and
% \IEEEauthorblockN{}
% \IEEEauthorblockA{\textit{School of Artificial Intelligence} \\
% \textit{Beijing University of Posts and Telecommunications}\\
% Beijing, China \\
% }
\and
\IEEEauthorblockN{Songpo Wang, Lei Yang}
\IEEEauthorblockA{
\textit{DeepScience Tech Ltd. }\\
Beijing, China\\
wangsongpo@deepscience.cn, yanglei@deepscience.cn}
% \and
% \IEEEauthorblockN{}
% \IEEEauthorblockA{
% \textit{DeepScience Tech Ltd. }\\
% Beijing, China\\
% }
}

\maketitle

\begin{abstract}
Audio driven talking head synthesis is a challenging task that attracts increasing attention in recent years. 
Although existing methods based on 2D landmarks or 3D face models can synthesize accurate lip synchronization and rhythmic head pose for arbitrary identity,
they still have limitations, such as the cut feeling in the mouth mapping and the lack of skin highlights. The morphed region is blurry compared to the surrounding face.
A Keypoint Based Enhancement (KPBE) method is proposed for audio driven free view talking head synthesis to improve the naturalness of the generated video. 
Firstly, existing methods were used as the backend to synthesize intermediate results. Then we used keypoint decomposition to extract video synthesis controlling parameters from the backend output and the source image. After that, the controlling parameters were composited to the source keypoints and the driving keypoints. A motion field based method was used to generate the final image from the keypoint representation.
With keypoint representation, we overcame the cut feeling in the mouth mapping and the lack of skin highlights.
Experiments show that our proposed enhancement method improved the quality of talking-head videos in terms of mean opinion score.
\end{abstract}


\section{Introduction}
In many applications, such as virtual reality, digital body, video conferencing, and visual dubbing, one-shot audio-driven talking head synthesis is an important component.
Early research relied on motion capture by art experts, and could only be used in film and games, 
which was labor-intensive and time-consuming \cite{edwardsJALIAnimatorcentricViseme2016}\cite{zhouVisemenetAudiodrivenAnimatorcentric2018b}. 
In recent years, significant progress has been made in this area, and a number of deep learning methods have been proposed \cite{taylorDeepLearningApproach2017}\cite{dasSpeechdrivenFacialAnimation2020}\cite{luLiveSpeechPortraits2021a}\cite{phamSpeechdriven3DFacial2017}\cite{thiesNeuralVoicePuppetry2020}\cite{jiAudiodrivenEmotionalVideo2021}\cite{guoAdnerfAudioDriven2021}\cite{prajwalLipSyncExpert2020a} in order to learn the warping from audio to expression. 
For example, Wav2lip \cite{prajwalLipSyncExpert2020a} uses a end-to-end framework, and synthesizes lower half of the face.
Many methods use 2D facial landmark \cite{chenHierarchicalCrossmodalTalking2019} \cite{suwajanakornSynthesizingObamaLearning2017a}\cite{dasSpeechdrivenFacialAnimation2020} or 3D head model \cite{andersonExpressiveTextdriven3D2013}\cite{thiesNeuralVoicePuppetry2020}\cite{songEverybodyTalkinLet2022}\cite{chenTalkingheadGenerationRhythmic2020}\cite{zhouMakeltTalkSpeakerawareTalkinghead2020}\cite{richardAudioandGazedrivenFacial2021}\cite{jiAudiodrivenEmotionalVideo2021} as a transit medium.
Because 2D facial landmark does not contain information about depth perception, most methods using 2D facial landmark \cite{luLiveSpeechPortraits2021a}\cite{suwajanakornSynthesizingObamaLearning2017a} do not support viewpoint editing.

In order to achieve more flexible manipulation, neural rendering methods are proposed. They are identity independent, and can change viewpoint in latent space.
Neural radiation fields (NeRF) \cite{guoAdnerfAudioDriven2021} is proposed to avoid additional intermediate representations.
However, it is still a difficult task to control the head pose and expression at the same time. To overcome this limitation, keypoint-based methods are proposed \cite{wangOneShotFreeViewNeural2021a}\cite{siarohinFirstOrderMotion2019}. They can render hair and sunglasses precisely that are not possible for 3D-based methods, and are able to change the viewpoint that are not possible for 2D-based methods.

To overcome the cut feeling in the mouth mapping and the lack of skin highlights, we propose a \textbf{Keypoint Based Enhancement (KPBE)} method for audio-driven free view talking head synthesis.
Our approach consists of a backend and a frontend. The backend is model-free. Using audio and source images as input, the existing backend methods synthesize intermediate results.
The frontend contains five modules: canonical keypoint estimator, appearance feature estimator, head pose and expression estimator, motion field estimator, and generator.
Canonical keypoint estimator is for estimating customized keypoints for the different images. 
Appearance feature estimator is for extracting the appearance features such as skin and color of eyes.
Head pose and expression estimator is for extracting head pose and expression from backend output. 
Specifically, head pose is determined by a rotation matrix and a translation vector. 
Expression is parameterized by vectors with the same number of the canonical keypoints.
Motion field estimator is for compositing the motion vectors in 3D space. Motion vector is obtained via pair of keypoints extracted from source image and driving image.
Generator is for generating final video from appearance feature and composited motion field.
Using this workflow we can get appearance features that maintain the skin highlights, and the generator avoids the cut feeling in the mouth mapping.
In addition, we can change the viewpoint by user-defined head pose matrix.

The contributions of our work are two folds:
\begin{itemize}
    \item A keypoint-based model-free enhancement method for audio-driven talking head synthesis is proposed, which can composite the head pose and the lip motion naturally.
    \item Head pose can be manipulated by user-defined rotation matrix and tranlation vector.
\end{itemize}
The rest of the paper is organized as follows. Section \ref{related works} introduces the related works about talking head synthesis. Section \ref{method} represents the architecture of our method and details of each part. Experiments and results are shown in Section \ref{experiments}. Section \ref{conclusions} concludes the paper. 

\begin{figure}[t]
  \centering
  \includegraphics[width=\linewidth]{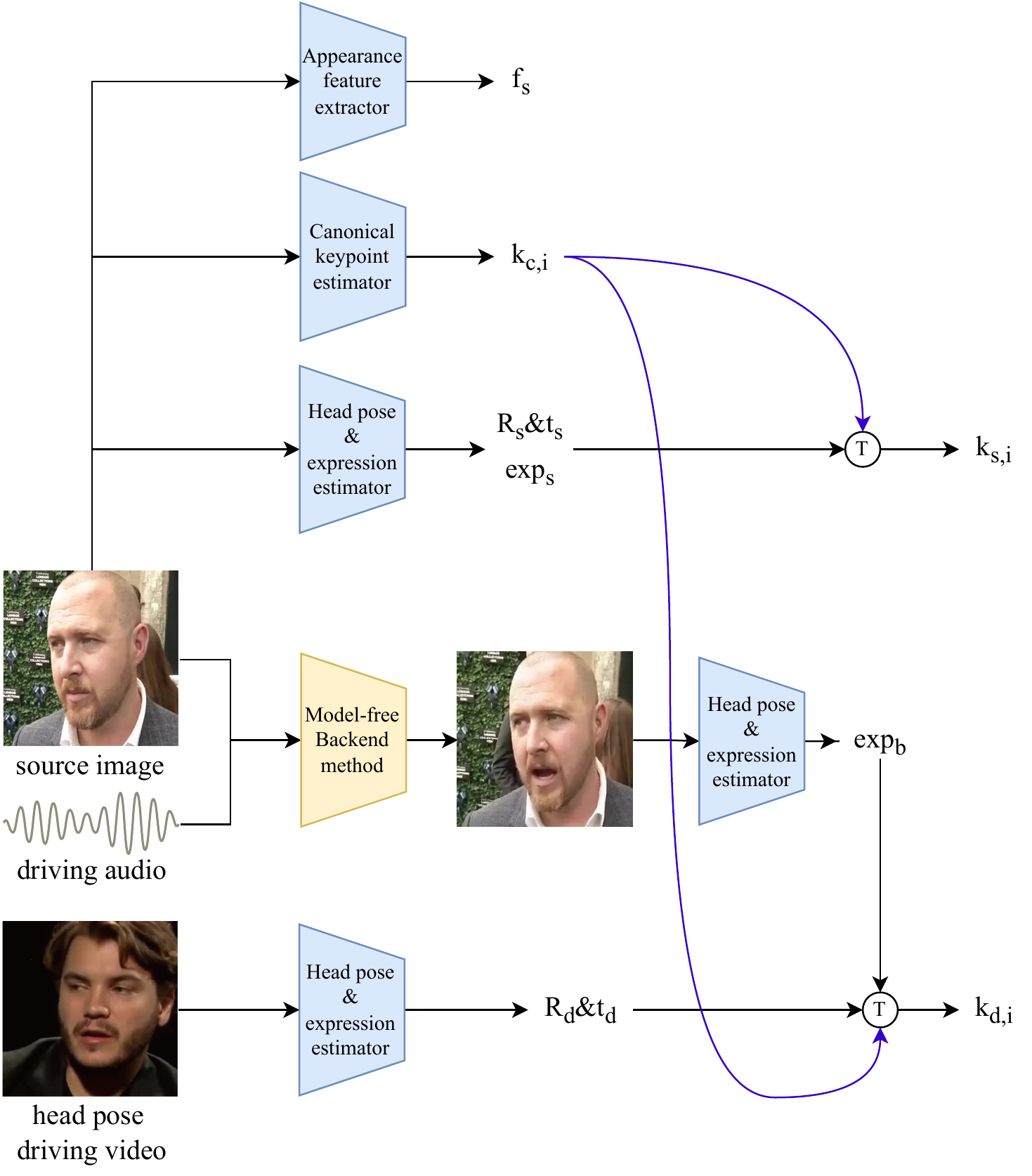}
  \caption{Keypoint decomposition. The appearance features, 3D canonical key points, head pose, and expression are extracted from the source image. Meanwhile, expression is extracted from backend output, and head pose is extracted from the head pose driving video. Applying the corresponding head pose and expression to the canonical keypoint, we get the source keypoint and the driving keypoint.}
  \label{fig:model1}
\end{figure}

\section{Related Works} \label{related works}

\textbf{Audio-driven talking head synthesis}.
Driving a talking head with audio is the task of synchronizing image frames of a video with arbitrary audio.
There are two lines of work: one is 3D-based and the other is 2D-based.
The early 3D-based methods attempt to build the relationship between audio features and lip motions by hand \cite{zhouVisemenetAudiodrivenAnimatorcentric2018b}\cite{andersonExpressiveTextdriven3D2013}, so they need an expert of the field. 
A well-known person-specific 3D-based work that does not rely on expert is   \cite{suwajanakornSynthesizingObamaLearning2017a}. The authors generated talking-head videos of president Obama, and focused on synthesizing natural head pose and accurate lip motion. However, the method needs a large number of videos of a specific person.
2D-based methods \cite{jamaludinYouSaidThat2019}\cite{prajwalLipSyncExpert2020a} can achieve identity independent expression synthesis by replacing part of the face. 
Recent 2D-based methods can also synthesize head pose using facial landmarks \cite{zhuArbitraryTalkingFace2021}\cite{zhouPoseControllableTalkingFace2021a}.
And some methods achieve real-time speed synthesis \cite{zakharovFastBilayerNeural2020a}.
Latest 3D-based methods can perform identity-independent synthesis\cite{chenTalkingheadGenerationRhythmic2020}. However, realistic hair cannot be generated , because it is difficult for a network to learn the model of these high poly models.
All these methods have only a limited capabilities for manipulating head pose and viewpoint.

\textbf{Neural rendering based talking head synthesis}.
Sitzmann \textit{et al.} \cite{sitzmann2019scene} used neural networks to represent the 3D shape or appearance of scenes, and they sampled point set in space to represent the appearance of an object.
A method was originally presented by Siarohin \textit{et al.} \cite{siarohinFirstOrderMotion2019} to get the warping between sparse keypoints and the motion fields from them.
Wang \textit{et al.} \cite{wangOneShotFreeViewNeural2021a} used 3D-based keypoint warping to overcome the shortcomings of the previous method. 
We use the similar idea for keypoint decomposition in our method.

Recently, Neural Radience Fields (NeRF) \cite{mildenhallNerfRepresentingScenes2020} has gained many achievements in neural rendering tasks. They transformed the 3D appearance features to ray sampling results of volume. AD-nerf \cite{guoAdnerfAudioDriven2021} applied this idea to talking head synthesis. Neural rendering methods achieve free view head pose synthesis, and have the ability to learn the depth information from the 2D image. We use a similar framework for extracting 3D appearance feature.

\section{Method} \label{method}

\begin{figure*}[t]
  \centering
  \includegraphics[width=\textwidth]{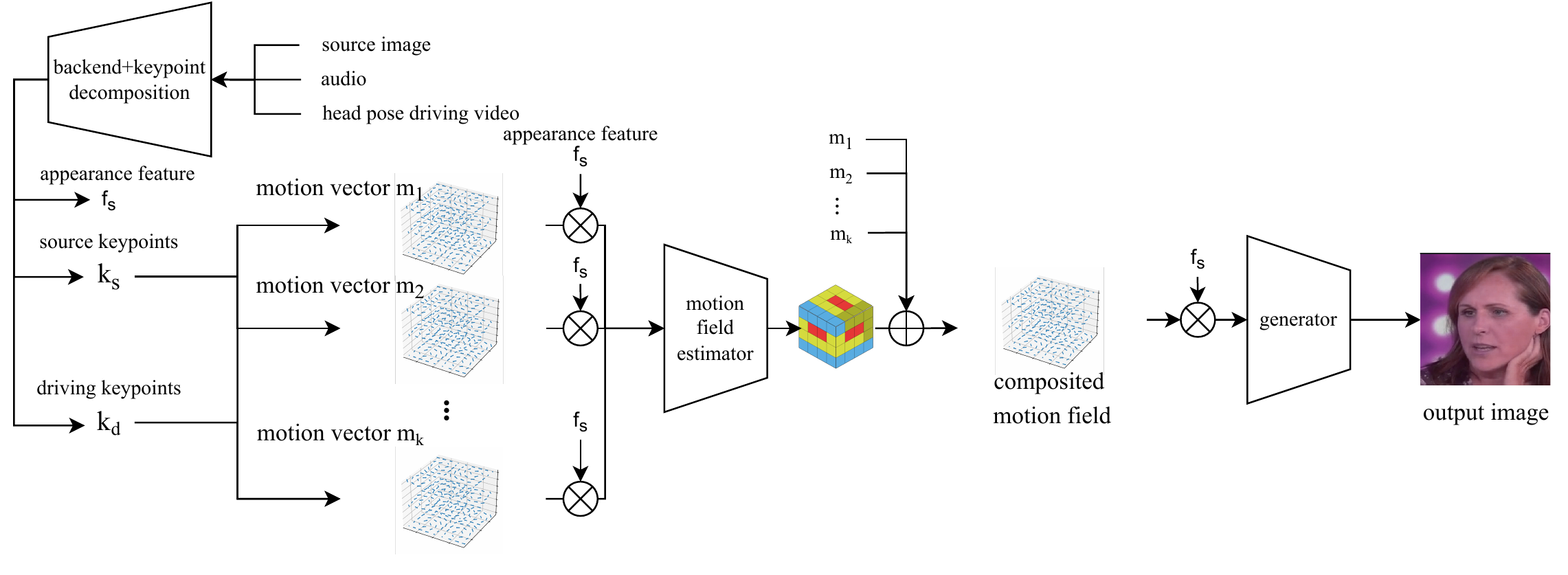}
  \caption{Generating from keypoint representation. Source keypoints and driving keypoints are derived from keypoint decomposition. We combined each pair keypoints, and used first order approximation to get the motion vectors. 
  The motion field mask is obtained by the motion field estimator.
  Then the mask is applied on the sum of motion vectors to get the composited motion field. The appearance feature is warped by composited motion field. After that the warped appearance feature is used to generate the final output image. }
  \label{fig:model2}
\end{figure*}

We proposed a \textbf{Keypoint Based Enhancement (KPBE)} method for audio-driven free view talking head synthesis.
The synthesis framework
consists of two steps: keypoint decomposition and generating from keypoint representation, as illustrated in Fig. \ref{fig:model1} and Fig. \ref{fig:model2}, respectively.
Specifically, the keypoint decomposition contains two parts. One is the backend which synthesizes the intermediate video from the audio. 
The other part belongs to the keypoint-based frontend which enhances the result of the backend output. Using keypoint decomposition, it extracts source keypoints and driving keypoints from the inputs. 
In the step of generating from keypoint representation part, we use the keypoints obtained by above part as input, and use the motion field based generator to synthesize the final image.

\subsection{Audio driven backend}
In this paper, we use PC-AVS and Wav2Lip as our backend methods (indicated by the yellow block in Fig. \ref{fig:model1}). 
Driving audio and the source image are used as the input of the backend model. The output of the backend is the intermediate video, and it contains the lip motion information. We extract head pose and expression from the video for enhancement step.

\subsection{Keypoint-based enhancement frontend}
The frontend contains five modules (indicated by the blue blocks in Fig. \ref{fig:model1}):  
Canonical keypoint estimator, appearance feature estimator, head pose and expression estimator, motion field estimator and generator.

\subsubsection{Canonical keypoint estimator}
Using the canonical keypoint estimator, we can extract canonical keypoint from the source image. Note that canonical keypoint is the keypoint associated only with the identity, and not affected by head pose and expression.
These extracted keypoints shall encode a person’s head
geometry feature in a neutral expression and pose.
Specifically, we extracted 15 3D canonical keypoints from the source image in this paper.

\subsubsection{Appearance feature estimator}
With the appearance feature estimator, the appearance features are extracted from the source image. Appearance feature are features related to the appearance of the face, such as skin and hair style. A 3D extraction network is applied to extract 3D appearance feature. It is important to use 3D feature. Only with the 3D appearance feature, can we restore it back to the photo after rotation and translation.

\subsubsection{Head pose and expression estimator}
The head pose and expression estimator is for estimating head pose and expression from the image. This module is used in two places in our approach.
One is for estimating the expression from the backend output, and the other is for estimating the head pose from the head pose driving video. 
Specifically, head pose contains rotation matrix and translation vector. Rotation matrix controls the 3D rotate angle of the head, and translation vector controls the center position of the head.
Expression is parameterized by vectors with the same number of the canonical keypoint.

With the canonical keypoint, head pose and expression, we give the equation of the 3D keypoint decomposition:
\begin{equation}
k_{s,i} = T(k_{c,i}, R_s, t_s, exp_{s}) = R_s k_{c,i} + t_s + exp_{s}
\label{skp}
\end{equation}
\begin{equation}
k_{d,i} = T(k_{c,i}, R_d, t_d, exp_{b}) = R_d k_{c,i} + t_d + exp_{b}
\label{dkp}
\end{equation}
where $k_{c,i}$ is the \textit{i}-th canonical keypoint extracted from the source image. $R_s$ and $t_s$ are the rotation matrix and the translation vector extracted from the source image. $exp_s$ is the expression extracted from the source image. $k_{s,i}$ is the \textit{i}-th source keypoint. $R_d$ and $t_d$ are the rotation matrix and the translation vector extracted from the head pose driving video. $exp_b$ is the expression extracted from the backend output. $k_{d,i}$ is the \textit{i}-th driving keypoint.
As illustrated in Fig. \ref{fig:model1}, the source keypoints are obtained by compositing head pose, expression and canonical keypoint extracted from the source image. We get driving keypoint by compositing head pose extracted from the head pose driving video, expression extracted from backend output and canonical keypoint extracted from the source image. Source keypoint and driving keypoint are uniquely determined by identity, expression and head pose.
It is of paramount importance for our approach to do the 3D keypoint decomposition. Our approach differs from prior 2D-based audio-driven talking head synthesis methods with regard to the keypoint decomposition. The keypoint decomposition helps learn controllable representations.
Note that unlike OSFV\cite{wangOneShotFreeViewNeural2021a}, our model is audio driven, and only extracts expression feature from the driving video. Our frontend step learns how to composite the feature extracted from different images, and restore the final image. With the keypoint decomposition, we can maximize the retention of appearance feature. This is helpful for reducing the cut feeling in the mouth mapping and the lack of skin highlights.

\subsubsection{Motion field estimator}
From the above three modules we can get source keypoints and driving keypoints. 
As shown in Fig. \ref{fig:model2}, the source keypoints $k_s$ and the driving keypoints $k_d$ are used as input to get motion vectors. Each pair of keypoints can be used to get a motion vector. A motion vector is a field that defines the movement tendency of every point in the 3D space. It is estimated by the same method of first-order approximation \cite{siarohinFirstOrderMotion2019}. We apply each motion vector to appearance feature and get the appearance feature warped by a motion vector. All warped appearance features are used as input for motion field estimator. The output is a 3D mask that weights different parts of the head. Then we apply the mask on the sum of the all motion vectors. The output is the composited motion field and a 2D occlusion mask, and we use this field to warp the appearance feature to get the warped appearance feature. 

\subsection{Generator}
Using the warped appearance feature obtained by the motion field estimator as input, the generator transforms the 3D warped appearance feature to the 2D final output image. Note that the warped appearance feature is multiplied by the occlusion mask after a convolution layer. For projecting the 3D feature to the 2D image, the framework of the generator uses a series of 2D upsampling layers. 
% The generator takes the warped 3D appearance features wpfsq and projects them back to 2D. Then, the
% features are multiplied with the occlusion mask o obtained
% from the motion field estimator M. Finally, we apply a series
% of 2D residual blocks and upsamplings layers to obtain the
% final image.

\subsection{Training}

Because we hope that the expression from the backend can be used in frontend,
 the loss of this step consists of perceptual loss and expression loss.
Loss functions are shown as follows:
\begin{equation}
    \mathcal{L}_p = \sum_{i \in Pyr} \Vert{VGG(Pyr_i G) - VGG(Pyr_i s)}\Vert_1
    % \mathcal{L}_p = \Vert{VGG(Pyr(G)) - VGG(Pyr(s))}\Vert_1
\end{equation}
\begin{equation}
    \mathcal{L}_{exp} = \Vert E_{d, i}\Vert_1
\end{equation}

$\mathcal{L}_p$ is the perceptual loss that helps to produce sharp-looking outputs.   VGG is a face recognition model\cite{omkar2015deep}. A pre-trained VGG framework is used to get the features of the final output image and the source image, and the L1 loss is computed between the features. 
$G$ is the generated final image. $s$ is the source image. 
We use a pyramid structure, $Pyr_i G$ and $Pyr_i s$ means the $i_th$ scaled images.
First features are extracted using the VGG network. 
After that, two images are fed into a downsample layer, then we use the VGG network to extract features from the sownsampled images and compute the L1 loss once again. We repeat this process five times, and compute L1 loss in different resolutions. Losses of different resolutions are added to compute the pyramid perceptual loss.

Expression loss $\mathcal{L}_{exp}$ is for penalizing the large keypoint deformation because the expression is a relatively small change compared to the head pose. Using the expression loss, we can clamp the magnitude of the keypoint deformation.

Our model is trained in two steps. First, we used the loss functions as the loss functions used in OSFV\cite{wangOneShotFreeViewNeural2021a} to train the frontend. 
Second, we added the backend to fine-tune the frontend. Our model was trained on Voxceleb1 which consists of talking head videos, and each video contains single person. The input of the second step is a source image sampled from video and a driving image sampled from backend output.

\section{Experiments} \label{experiments}
\subsection{Dataset}
Our experiment was based on Voxceleb1 \cite{nagraniVoxCelebLargeScaleSpeaker2017}.
We used the same preprocessing method in FOMM \cite{siarohinFirstOrderMotion2019}, 
which has 18672 videos for train and 525 videos for test.
Voxceleb1 had different resolution videos and all videos were converted to 25 fps in our training. 
Since there were many extreme head pose cases, different light conditions, and different ethnicity in the dataset, our model was robust.

\subsection{Implementation detail}
Canonical keypoint estimator, appearance feature estimator, head pose and expression estimator, motion field estimator and generator are with the same structure of the OSFV \cite{wangOneShotFreeViewNeural2021a}. We fine tuned the pretrained backend and frontend model with our training loss.

We used the ADAM optimizer with $\beta_1=0.5$ and $\beta_2=0.999$ and trained on 4 NVIDIA 24GB GTX3090 GPUs. The model was trained for 200 epochs, and each epoch needs 2 hours.

\subsection{Metrics}
We used PSNR and SSIM \cite{wang2004image} as metrics to quantify the faithfulness of the synthesized videos. 
SSIM measures the structural similarity between ground truth and synthesized videos. Compared with PSNR, it is more robust to global illumination changes.
Given a reference image $s$ and a test image $g$, both of which have a size of M×N, the PSNR
between s and g is defined by: 
$$
\operatorname{PSNR}(s, g)=10 \log _{10}\left(255^{2} / M S E(s, g)\right)
$$
where
$$
\operatorname{MSE}(s, g)=\frac{1}{M N} \sum_{i=1}^{M} \sum_{j=1}^{N}\left(s_{i j}-g_{i j}\right)^{2}
$$
PSNR is used to measure image reconstruction quality. It computes the mean squared error between the input and the output images.
SSIM is defined as:
$$
\operatorname{SSIM}(s, g)=l(s, g) c(s, g) s(s, g)
$$
where
\begin{equation}
\left\{\begin{array}{l}
l(s, g)=\frac{2 \mu_{s} \mu_{g}+C_{1}}{\mu_{s}^{2}+\mu_{g}^{2}+C_{1}} \\
c(s, g)=\frac{2 \sigma_{s} \sigma_{g}+C_{2}}{\sigma_{s}^{2}+\sigma_{g}^{2}+C_{2}} \\
s(s, g)=\frac{\sigma_{s g}+C_{3}}{\sigma_{s} \sigma_{g}+C_{3}}
\end{array}\right.
\end{equation}
$\mu_{s}$ and $\mu_{g}$ are the mean luminance of the images. $\sigma_{s}$ and $\sigma_{g}$ are the variance of the luminance. $\sigma_{s g}$ is the covariance between the two images s and g. The values of SSIM are from 0 to 1. Value 0 means no relation between images, and 1 means two images are the same image.

The test videos for our experiment were generated by random choosing video from the dataset as the ground truth. We randomly chose one frame from it as the source image, and use the same video as the head pose driving video.
We calculate these two metrics on the test set, and evaluate the mean score.

In addition, we also conducted a subjective test to evaluate the performance of our proposed method where subjects were recruited to rate the
mean opinion scores (MOS) of lip synchronization, head pose naturalness, and video realness. The value ranges from 1 to 5 (higher is better).

\subsection{Objective evaluation}
We chose Wav2lip \cite{prajwalLipSyncExpert2020a} and PC-AVS \cite{zhouPoseControllableTalkingFace2021a} as the backend method,
and compared our method with the results before enhancement for quantitative evaluation. As shown in Table \ref{tab:result_1}, both methods have improvement on SSIM and PSNR.
As illustrated in Fig. \ref{fig:slides}, we can see that our method has a better color balance than the PC-AVS. The PC-AVS results had a pale skin, and our method results had a normal skin. 
Wav2lip has a block of blurring arround the mouth,  and our method result does not have the blurring.

\subsection{Subjective  evaluation}
\begin{figure}[htbp]
  \centering
  \includegraphics[width=\linewidth]{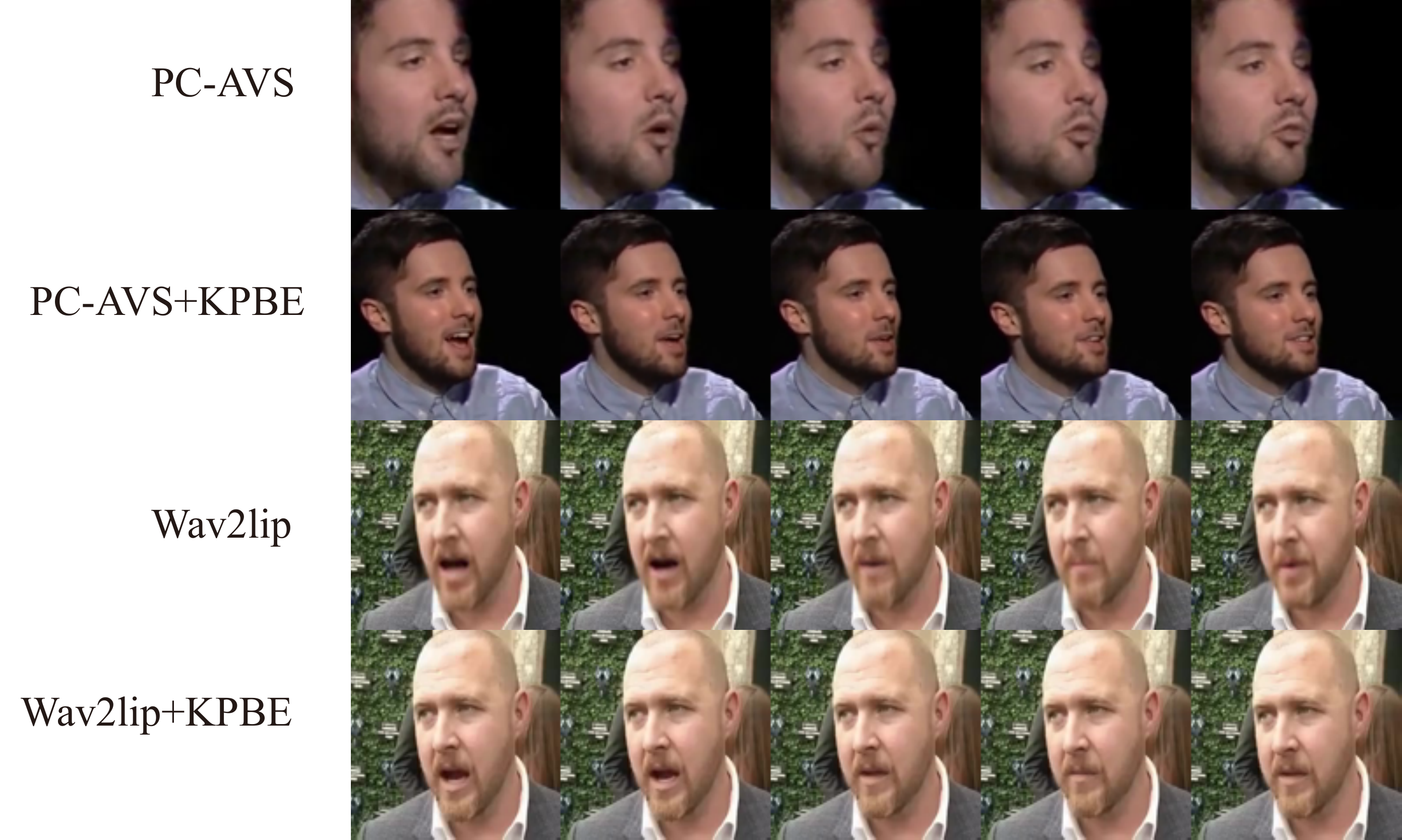}
  \caption{Synthesized video frames by PC-AVS, Wav2lip, and our method.}
  \label{fig:slides}
\end{figure}

We conducted a subjective test to compare different methods.
Each method synthesized 13 videos which were then used as stimuli. The subjective test was performed online where ten subjects were recruited to rate MOS of lip synchronization, head pose naturalness, and video realness. The subjects were asked to wear headphones and complete the experiment in a quiet environment.
% We measured the methods in three aspects. 
% Each method synthesized 13 videos which were then used as stimuli in the subjective test.
% Ten subjects were recruited to rate the mean opinion scores (MOS) of lip synchronization, head pose naturalness, and video realness.
% Subjects used headphones and conducted the experiment online in a quiet environment. 
Table \ref{tab:result_2} shows that with KPBE enhancement, the lip synchronization, head pose naturalness, and video realness of the enhanced videos are improved.
Fig. \ref{fig:slides} shows synthesized video frames by PC-AVS, Wav2lip, and their KPBE enhancements. 
First line and second line are the results of the PC-AVS and our method.
PC-AVS could synthesize pretty good head pose and lip motion, but they sometimes synthesized people without highlights on skin. When the head had a relatively large rotation, the face would be deficient. In contrast our model had a better face highlights and had no facial deficient.

The significant enhancement is on the Wav2lip method, and their method often synthesizes accurate lip motion but with low visual quality on the mouth part. Our method makes it overcome this shortcoming. From the third line and forth line in Fig. \ref{fig:slides}, we can see that the result of Wav2lip has some blurring at the mouth part. There is a dividing line between the neck and the face. Our method result has a clearer mouth, and there is no dividing line.

\begin{figure}[htbp]
  \centering
  \includegraphics[width=\linewidth]{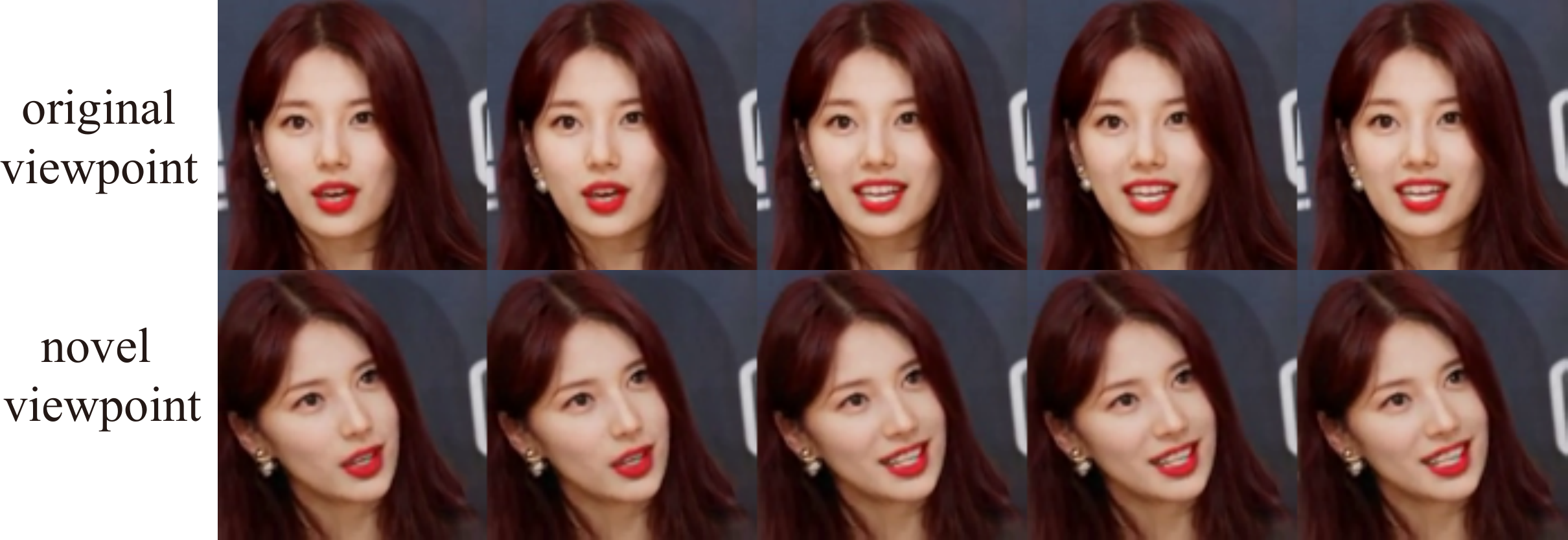}
  \caption{Novel viewpoint video synthesis. With the user-defined head pose matrix, we can reenact the audio-driven talking head video from a novel viewpoint.}
  \label{fig:viewpoint}
\end{figure}

\begin{table}[htbp]
  \caption{Objective results. We calculated the metrics on the test set, and
evaluate the mean score.}
  \label{tab:result_1}
  \centering
  \begin{tabular}{lll}
    \toprule
    \textbf{Method}      & \textbf{SSIM$\uparrow$}     & \textbf{PSNR$\uparrow$}   \\
    \midrule
    Wav2lip                   & 0.85        & 28.67         \\
    Wav2lip+KPBE(Ours)        & \textbf{0.90}        & \textbf{30.37}         \\
    PC-AVS                    & 0.80        & 22.19         \\
    PC-AVS+KPBE(Ours)         & \textbf{0.82}        & \textbf{23.64}         \\
    
    \bottomrule
  \end{tabular}
\end{table}

\begin{table}[!htbp]
  \caption{Mean Opinion Scores.  Value from 1 to 5, higher is
better.}
  \label{tab:result_2}
  \centering
  \begin{tabular}{llll}
    \toprule
    \textbf{Method}      & 
    \begin{tabular}{@{}c@{}}\textbf{Lip} \\ \textbf{Sync}\end{tabular} &
    \begin{tabular}{@{}c@{}}\textbf{Head Pose} \\ \textbf{Naturalness}\end{tabular} &
    \begin{tabular}{@{}c@{}}\textbf{Video} \\ \textbf{Realness}\end{tabular}\\
    \midrule
    Wav2lip \cite{prajwalLipSyncExpert2020a}                   
                                    & 4.11       & 4.03   &4.09         \\
    Wav2lip+KPBE(Ours)              & \textbf{4.14}       & \textbf{4.14}   & \textbf{4.27}         \\
    PC-AVS \cite{zhouPoseControllableTalkingFace2021a}                    
                                    & 3.44       & 3.12   &3.25         \\
    PC-AVS+KPBE(Ours)               & \textbf{3.70}       & \textbf{3.31}   &\textbf{3.41}         \\
    Ground\_truth                   & 4.59       & 4.50   & 4.73        \\
    
    \bottomrule
  \end{tabular}
\end{table}

\subsection{Novel viewpoint video synthesis}
To test the free-view video generation capabilities of our model, we synthesized the video from a novel viewpoint with the user-defined rotation matrix and translation matrix. The results are shown in Fig. \ref{fig:viewpoint}. Using the head pose matrix, we can freely control the rotation of the head pose. Using the translation matrix, we can control the displacement of the head.

\section{Conclusions} \label{conclusions}

Audio-driven talking head synthesis attracts increasing attention in recent years. 
We proposed a novel keypoint-based enhancement method for audio-driven talking head synthesis to generate enhanced video with better lighting balance and more natural expressions and head poses.
Using keypoint decomposition, our method could disentangle the image features into appearance features, canonical keypoints, and head pose matrix.
We extracted expression from the backend output and head pose from the head pose driving video. After that, the motion field based generator was used to generate the final image.
Experiments results verified that our method can reduce the cut feeling in the mouth mapping and the lack of skin highlights. Limited by the low resolution and language propensity of the dataset, our model cannot synthesize high-resolution videos and accurate lip movements in some languages. In the future, we will focus on improving the resolution and cross-language accuracy in audio-driven talking head synthesis.

\section*{Acknowledgements}

This work is supported by the Open Project Program of the National Laboratory of Pattern Recognition (NLPR) (202200042) and New Talent Project of Beijing University of Posts and Telecommunications (2021RC37).

% \nocite{*}

\vspace{12pt}

\end{document}